\documentclass[nohyperref]{article}

\usepackage{microtype}
\usepackage{graphicx}
\usepackage{subfigure}
\usepackage{booktabs} 

\usepackage{hyperref}

\usepackage[accepted]{icml2022}

\usepackage{amsmath}
\usepackage{amssymb}
\usepackage{mathtools}
\usepackage{amsthm}
\usepackage{color}
\usepackage{stmaryrd}

\usepackage[capitalize,noabbrev]{cleveref}

\theoremstyle{plain}

\theoremstyle{definition}

\theoremstyle{remark}

\usepackage[textsize=tiny]{todonotes}

\icmltitlerunning{PoF: Post-Training of Feature Extractor for Improving Generalization}

\usepackage{url}

\newcommand{\etal}{\textit{et al}.}
\newcommand{\etc}{\textit{etc}.}
\newcommand{\ie}{\textit{i}.\textit{e}.}

\newcommand{\argmin}{\mathop{\rm arg~min}\limits}

\begin{document}

\twocolumn[
\icmltitle{PoF: Post-Training of Feature Extractor for Improving Generalization}



\icmlsetsymbol{equal}{*}

\begin{icmlauthorlist}
\icmlauthor{Ikuro Sato}{equal,university,company}
\icmlauthor{Ryota Yamada}{equal,university}
\icmlauthor{Masayuki Tanaka}{university}
\icmlauthor{Nakamasa Inoue}{university}
\icmlauthor{Rei Kawakami}{university,company}
\end{icmlauthorlist}

\icmlaffiliation{university}{School of Computing, Tokyo Institute of Technology, Japan}
\icmlaffiliation{company}{Denso IT Laboratory, inc., Japan}

\icmlcorrespondingauthor{Ikuro Sato}{isato@c.titech.ac.jp}

\icmlkeywords{Machine Learning, ICML}

\vskip 0.3in
]



\printAffiliationsAndNotice{\icmlEqualContribution} 

\begin{abstract}
It has been intensively investigated that the local shape, especially flatness, of the loss landscape near a minimum plays an important role for generalization of deep models.  We developed a training algorithm called PoF: Post-Training of Feature Extractor that updates the feature extractor part of an already-trained deep model to search a flatter minimum. The characteristics are two-fold: 1) Feature extractor is trained under parameter perturbations in the higher-layer parameter space, based on observations that suggest flattening higher-layer parameter space, and 2) the perturbation range is determined in a data-driven manner aiming to reduce a part of test loss caused by the positive loss curvature. We provide a theoretical analysis that shows the proposed algorithm implicitly reduces the target Hessian components as well as the loss. Experimental results show that PoF improved model performance against baseline methods on both CIFAR-10 and CIFAR-100 datasets for only 10-epoch post-training, and on SVHN dataset for 50-epoch post-training.
Source code is available at: \url{https://github.com/DensoITLab/PoF-v1}.
\end{abstract}

\section{Introduction}
\label{sec_intro}

\begin{figure}[t]
{\normalsize
    \begin{center}
    {\tabcolsep=2mm
    \begin{tabular}{cc}
    \includegraphics[width=33mm]{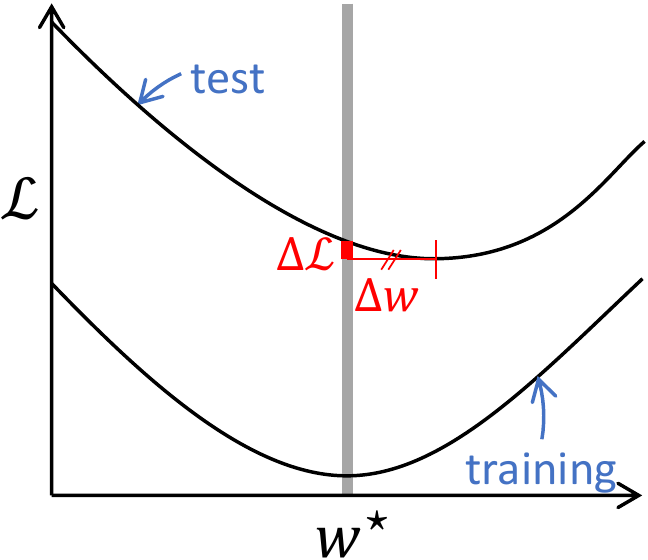}& 
    \includegraphics[width=33mm]{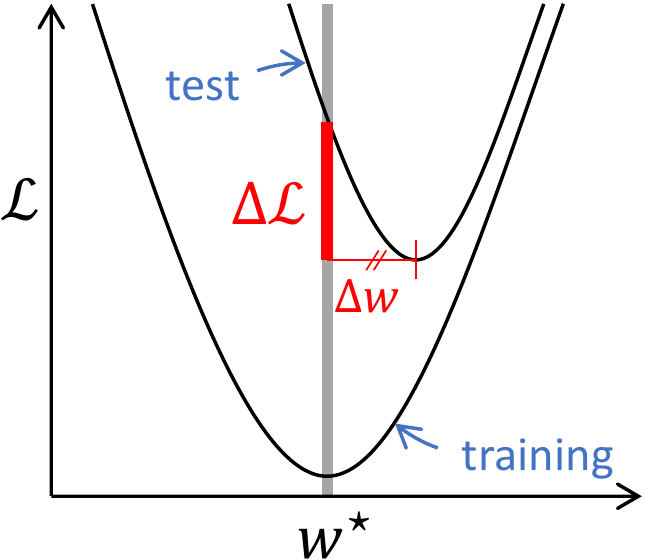} \\
    (a) Flatter minima & (b) Sharper minima \\
    \addlinespace[2.5mm]
    \includegraphics[height=31mm]{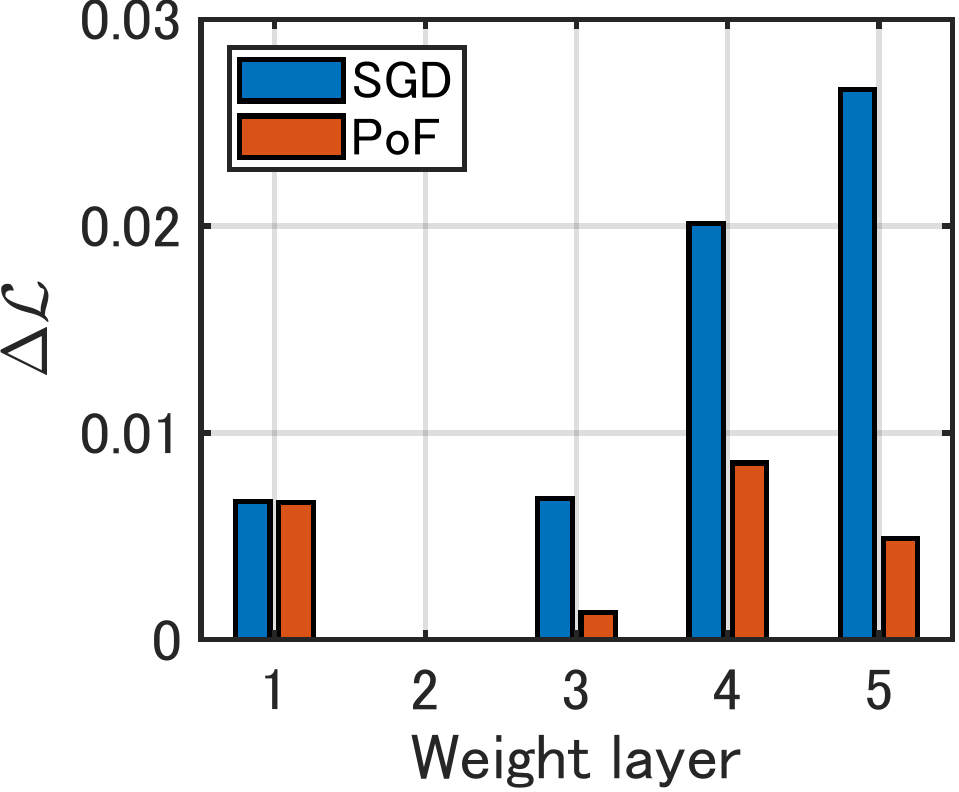}&
    \includegraphics[height=31mm]{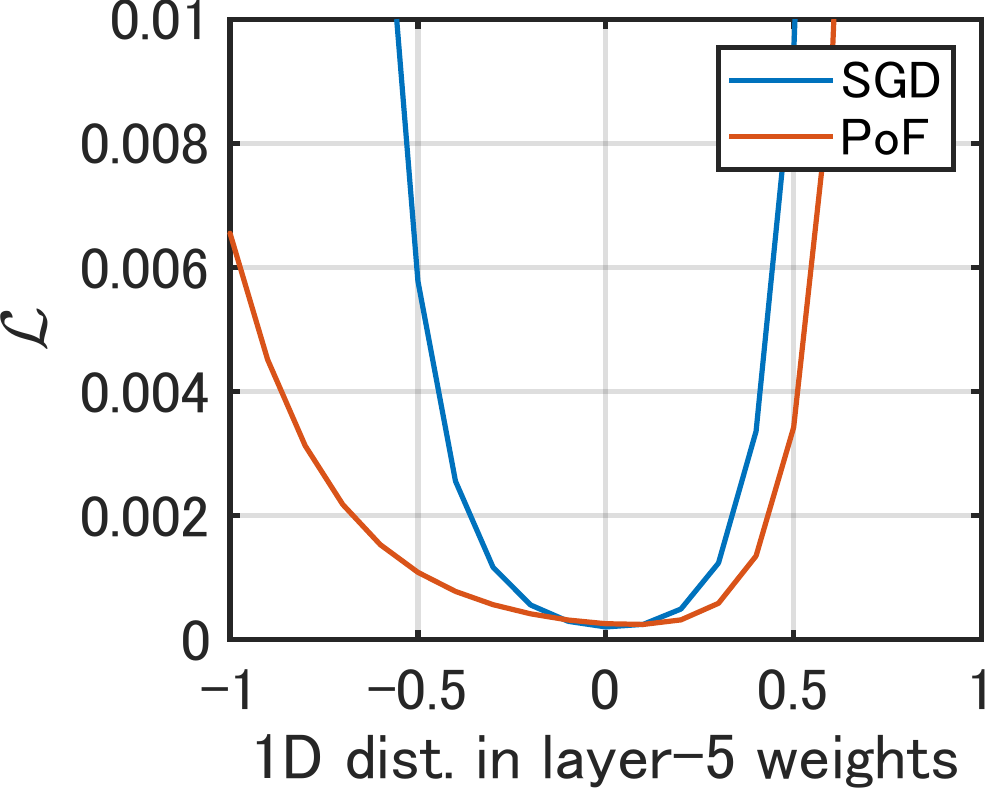}\\
    (c) $\Delta \mathcal{L}$ & (d) Loss landscapes
    \end{tabular}
    }
    \end{center}
}
    \caption{Sketches of local structures around minima (top), and behaviors of training methods with toy data (bottom). (a) Curvature-based loss increment $\Delta\mathcal{L}$ tends to be small for a relatively flat test-loss landscape. (b) $\Delta\mathcal{L}$ tends to be large for a sharper case. (c) Loss increments are evaluated for different layers of an MLP with 5 weight layers after a vanilla SGD and the proposed PoF. PoF successfully reduces $\Delta\mathcal{L}$ at the final layer. (d) Local loss landscapes of the training set along the principal eigenvectors of the Hessian matrices at the final layer. PoF expands the characteristic scale of the flat region.}
    \label{fig_explanation}
\end{figure}

It has been intensively discussed what conditions make deep models generalized for given datasets and network architectures.
Factors that affect learning dynamics, such as optimizers~\cite{10.5555/3491440.3491892, Keskar2017ImprovingGP, AshiaWilson, NEURIPS2020_08fb104b}, batch sizes~\cite{chaudhari2019entropy, paper:keskar17} and learning rate~\cite{chaudhari2018stochastic, goyal2017accurate} \etc, are known to affect generalization abilities.
Related to the learning dynamics, information theoretical aspects such as loss landscapes near a local minimum brought insights to the way a model acquire generalization ability.
Studies about the loss landscape~\cite{paper:Hochreiter97, paper:keskar17, paper:Jiang20, PAC_flat, paper:Jiang20} argue that a flatter local structure around a minimum (Fig.~\ref{fig_explanation} (a)) is preferred to a sharper one (Fig.~\ref{fig_explanation} (b)).
This argument holds sufficiently if distance $\Delta w$ between the training and test minimizers are the same in the flatter and sharper cases, as depicted in Fig.~\ref{fig_explanation} (a) and (b).

Recent optimization methods that seek flatter minima have been exhibiting to improve generalization of deep models for various tasks~\cite{paper:sam, paper:Jungmin21, paper:Donxian, paper:AMP, paper:Izmailov}.
While performance gain is likely obtained by these methods, 
practitioners need to examine such methods under different hyperparameter settings or combinations to figure out the best performing one for given dataset and network architecture.
Even if a best performing model is obtained after trials, no easy way is known to check if there is still a room to improve the performance.
Isn't there any strategy to examine an already-trained model to see if further performance improvement is possible, for instance, by actively seeking an even wider basin?

A technical question to implement this strategy would be in what subspace of the parameter space the flatness indicator should be improved.
Importance of flatness may vary for the lower and higher layer parameters.
In general, compared to the distribution of the higher layer features, the distribution of the lower layer features is not well separated; therefore, it is likely that the shift of feature distributions between the training and test sets is comparatively moderate in the lower layers.

In Fig.~\ref{fig_explanation} (c), we examined how much the test loss increases, denote by $\Delta\mathcal{L}$, from its minimum value near the training-set minimizer on an artificially generated toy dataset using a 5-layer MLP.
We extract the eigenvector corresponding the maximum eigenvalue of the Hessian block that is the second derivative of the loss function with respect to the parameters at each layer.
Then we measure $\Delta\mathcal{L}$ along each of the eigenvectors, plotted in Fig.~\ref{fig_explanation} (c).
Typically, but not always, higher layers tend to indicate larger $\Delta\mathcal{L}$ than lower layers.
This means that there likely exists a better solution in the vicinity of the current solution along the direction that has the largest curvature in the higher layer parameter space.
Enhancing flatness along such a direction would be more efficient, rather than arbitrarily chosen directions.

Another technical question would be in what range the loss landscape should become flat.
Suppose that the training loss landscape becomes fairly flat in a certain region around the minimum.
If the gap between the training and test loss minimizers, denoted as $\Delta w$, is larger than the characteristic scale of the flat region (see Fig.~\ref{fig_explanation} (b)), expanding the flat region would improve generalization performance.
In contrast, if such a gap is similar or less than the characteristic scale of the flat region (see Fig.~\ref{fig_explanation} (a)), further expansion of the flat region would have little effect and simply reducing (the zero-th derivative of) the loss would be a better approach.

In this paper, we propose a training method called PoF: Post-Training of Feature Extractor that updates the feature extractor part of an already-trained deep model to search a flatter minimum for improving generalization.
Our method addresses the abovementioned technical issues.
Let us arbitrarily divide a deep model into two parts: the feature extractor and the classifier.
For a case of 2D convolutional neural network, the former may include the local feature processing layers and a layer that convert the local features to the global features, such as the global average pooling layer~\cite{Lin2013network}, and the latter includes all the subsequent layer(s).
We summarize our main contributions below.
\begin{itemize}
    \item The proposed training method, PoF, post-trains the feature extractor part of a given deep model whose parameters are already at a local minimum  by some method.
    PoF provides a practical means for searching a better-performing model, given that the computational time required by PoF is shorter than a typical end-to-end training from random initialization.
    \item PoF is designed to flatten the local shape of a loss function near a minimum in the classifier parameter space by gradually changing feature-extractor parameters, based on an assumption that flattening the loss landscape in the classifier parameter space enhances robustness, similar to the concept of maximum-margin classifiers.
    \item The characteristic range where PoF enhances flatness is determined in a data-driven manner to balance the 0-th and 2nd order derivative terms so that loss increment caused by non-zero curvature is well reduced. 
    No hand tuning is required to set an upper-bound of the range.
\end{itemize}

\section{Related Work}

\subsection{Flatness and Generalization}

The relationship between the local loss landscape and the generalization ability of a minimum has been discussed extensively in theoretical and empirical literature \cite{paper:Hochreiter97, paper:keskar17, PAC_flat, paper:Jiang20, paper:dinh17}. 
The previous section described an intuitive picture explaining why a flatter minimum likely generalizes.
Another view is given by the bits-back argument~\cite{paper:Hinton93, paper:Honkela04}.
It states models that are stable against weight perturbations can be described with fewer bits.
According to the minimum description length (MDL)~\cite{paper:Rissanen78} or similar criteria, models that can be represented with a smaller number of bits are expected to have better generalization abilities.
Both the geometrical and description-length points of views suggest flatter minima are preferable for generalization.

With deep neural networks, there are a number of studies that aim to measure the flatness of the loss landscape.
For example, \citet{paper:keskar17} measured the flatness by the worst loss around the minima.
\citet{NEURIPS2018_a41b3bb3} visualizes loss landscapes with findings that sharp minimizers tend to have larger generalization error.
Loss landscape visualization was carried out with minimization trajectories~\cite{Goodfellow2015} and with parameter interpolations~\cite{DBLP:journals/corr/ImTB16}.

Hessian matrices are sometimes utilized to quantify flatness with eigenvalues~\cite{paper:Sagun16, paper:Wu17, paper:Zhang18}
and spectral norm~\cite{paper:Yao18}.
However, measuring the flatness is still an open problem in general due to high dimensionality and architectural complexity of deep networks.

\subsection{Optimization Methods and Flatness}

For a given network architecture, different optimization methods tend to reach solutions that are different in terms of flatness.
Stochastic gradient descent (SGD) is known to be biased towards flat minima \cite{paper:Jastrzebski18, paper:Maddox20}.
The stochastic noise in the parameter updates leads to parameter convergence in distribution with similar loss from a Langevin-dynamics based modeling~\cite{chaudhari2018stochastic}.
Stochastic Weight Averaging (SWA) was proposed to find a flatter point near a minimizer reached by SGD, but SWA itself does not work like an optimizer that can find a wider basin. 
Recently, optimization algorithms that seek flatter minima have attracted attention.
The sharpness-aware minimization (SAM) \cite{paper:sam} based on the PAC-Bayes generalization bound \cite{paper:Becker03} has been shown to be effective for various image classification tasks.
There are some extensions of SAM including adaptive optimization methods \cite{paper:Chen20, paper:Jungmin21}.
These sharpness-aware methods apply parameter perturbations within some radius at each iteration, aiming to make the perturbed region flat.
PoF also adopts a parameter perturbation, whose range is determined in data-driven manner to reduce a specially designed effective loss as described in the next section.

Curvature estimation of a loss surface by approximated Hessian matrices or Fisher information matrices~\cite{paper:Roux, paper:Botev17, paper:Grosse16, paper:Martens15, paper:Pauloski21} is related to the flatness-based optimization.
A second-order term appears in the formulation of SAM, but it is simply dropped to reduce the computational cost from a practical perspective.
PoF is designed to implicitly reduce some Hessian components without a need to directly compute Hessian or Fisher information matrices.




\subsection{Co-Adaptation Prevention between Layers}

Some previous work discussed co-adaptation prevention between layers, in particular, a feature extractor and a classifier.
Prevention of co-adaptation among neurons likely brings a positive effect on generalization~\cite{Hinton2012}.
FOCA~\cite{paper:foca} avoids between-layer co-adaptation by using many random weak classifiers during optimization.
\citet*{paper:Hojjat20} proposed a method to adaptively assign dropout rate according to the co-adaption pressure.
\citet{paper:Wei20} makes a series of weak classifiers to decouple co-adaptation.
Our work can be viewed as a method to weaken co-adaptation between the feature extractor and the classifier.



\section{Post-Training of Feature Extractor}

This section explains the proposed method PoF that post-trains a feature extractor based on a specially-designed flatness index.
In the following, we consider a supervised setting.
Let $(x, t)$ be a data sample consisting of a real-valued input data $x$ and the corresponding real/integer-valued target data $t$, respectively.
The training dataset $\mathcal{D}$ contains $n_\mathcal{D}$ such training samples.
We denote a feature extractor as a function of $x$ by $F_\phi(x)$ with the feature-extractor parameter set $\phi$.
Similarly, a classifier is denoted as a function of feature by $C_\theta (F_\phi(x))$ with the parameter set $\theta$.
The loss function of the training dataset is given by
\begin{equation}
    \mathcal{L}_\mathcal{D}(\phi,\theta) = \frac{1}{n_\mathcal{D}}\sum_{(x,t)\in\mathcal{D}} L(C_\theta (F_\phi(x)), t),
\end{equation}
where $L(\cdot)$ is a sample-wise loss function such as squared error or cross entropy.
Similar to $\mathcal{L}_\mathcal{D}(\phi,\theta)$, we denote a mini-batch loss by $\mathcal{L}_\mathcal{B}(\phi,\theta)$, which is an averaged sample-wise loss within a given mini-batch $\mathcal{B}$ of size $n_\mathcal{B}$.
Let $(\phi_0, \theta_0)$ be a pair of parameter sets that are given by some training method so that the training loss $\mathcal{L}_\mathcal{D}(\phi, \theta)$ is regarded to be locally minimized at $(\phi, \theta) = (\phi_0, \theta_0)$.

We assume that the loss function is locally convex around the local minimum.
In general, the number of training samples is finite, so the minimizer of training loss, $(\phi_0, \theta_0)$, does not exactly coincide with the closest minimizer of the test loss, $(\phi_\mathrm{test}, \theta_\mathrm{test})$.
As illustrated in Fig.~\ref{fig_explanation}, how large the characteristic scale of the flat region
is compared to the parameter distance between $(\phi_0, \theta_0)$ and $(\phi_\mathrm{test}, \theta_\mathrm{test})$ is important for generalization.
Importance of flatness of a loss landscape has been pointed out; however, extending a flat region much beyond this parameter distance would be meaningless.
In this case, naively descending the loss may be more effective.
PoF is designed not just to make a loss landscape flat within a predefined region, but to control the characteristic scale of the flat region from the abovementioned perspective in a data-dependent fashion.

In general, it is impossible to control the flatness or Hessian components for a \textit{test set}.
In this work, we simply assume that the shape / curvature of the test loss landscape is interlocked with that of the training loss landscape, but the positions of their minima can somewhat differ.
Such a condition was also implicitly assumed in \cite{paper:Hochreiter97, paper:keskar17}.

\subsection{Algorithm}

\begin{algorithm}[t]
\caption{Post-training of Feature Extractor (PoF)}
\label{alg1}
\begin{algorithmic}
{\small
\STATE {\bf Input}: feature-extractor parameters $\phi_{0}$;
    classifier parameters $\theta_{0}$; training data $\mathcal{D}$;
    expansion factor $\gamma$
\vspace{1.0mm}
\STATE $\phi = \phi_{0}$
\FOR{$t=1:T$}
    \STATE Draw a mini-batch $\mathcal{B}$ from $\mathcal{D}$.
    \STATE Obtain $\xi_\mathcal{B}^\star$ by Eq.~(\ref{eq_xistar}).
    \STATE Draw another mini-batch $\tilde{\mathcal{B}}$ from $\mathcal{D}$.
    \STATE Obtain $\Delta \phi$ by Eq.(\ref{eq_DeltaPhi}) with $\gamma$.
    \STATE $\phi \leftarrow \phi + \Delta \phi$
\ENDFOR
\vspace{1.0mm}
\STATE {\bf Output}: feature-extractor parameters $\phi^\star = \phi$
}
\end{algorithmic}
\end{algorithm}

\begin{figure}[t]
{\normalsize
    \begin{center}
    \includegraphics[height=55mm]{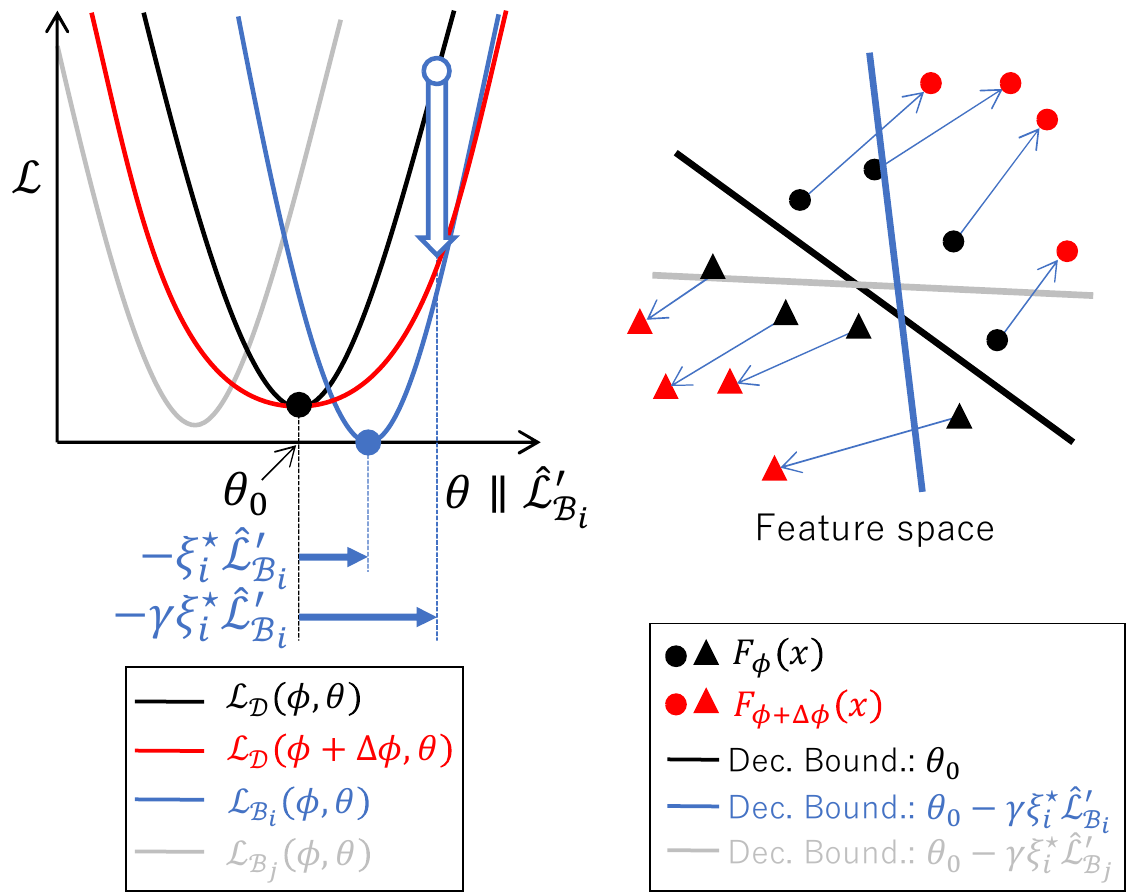}
    \end{center}
}
    \caption{Illustration of how loss landscape becomes flat (left) and the allowed margin around the decision boundary changes (right) by PoF. Left: The algorithm starts from the minimum (black dot) to search for a minimum of a mini-batch loss in the classifier parameter space (blue dot). The expansion factor $\gamma$ takes the point to another point (blue circle). Feature-extractor parameter updates with perturbed classifiers eventually yield a flatter loss landscape (red curve). Right: Feature distributions change under feature-extractor parameter updates with perturbed decision boundaries so that more margin is allowed around $\theta_0$-classifier.}
    \label{fig_algorithm}
\end{figure}

PoF is a perturbation-based method to seek a flatter minimum, \ie, loss gradients are computed with respect to shifted parameters at each iteration.
As illustrated in Fig.~\ref{fig_explanation} (c), we assume that the deterioration in loss caused by the positional discrepancy between the training and test minimizers is most severe in the classifier parameter space.
PoF aims to update the feature extractor so that the returned features yield a flat loss landscape to an appropriate extent in the classifier parameter space.
Making the classifier parameters or the corresponding decision boundary ``loose" is analogous to the concept of the maximum margin method (see Fig.~\ref{fig_algorithm}).

\begin{figure}[t]
{\normalsize
    \begin{center}
    \vspace{5mm}
    {\tabcolsep=1mm
    \begin{tabular}{cc}
    \includegraphics[height=34mm]{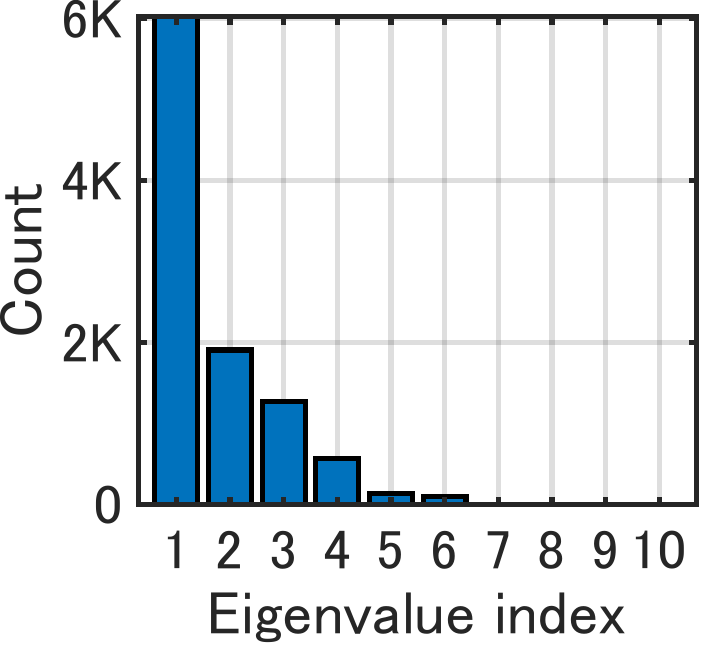}&
    \hspace{0mm}
    \includegraphics[height=34mm]{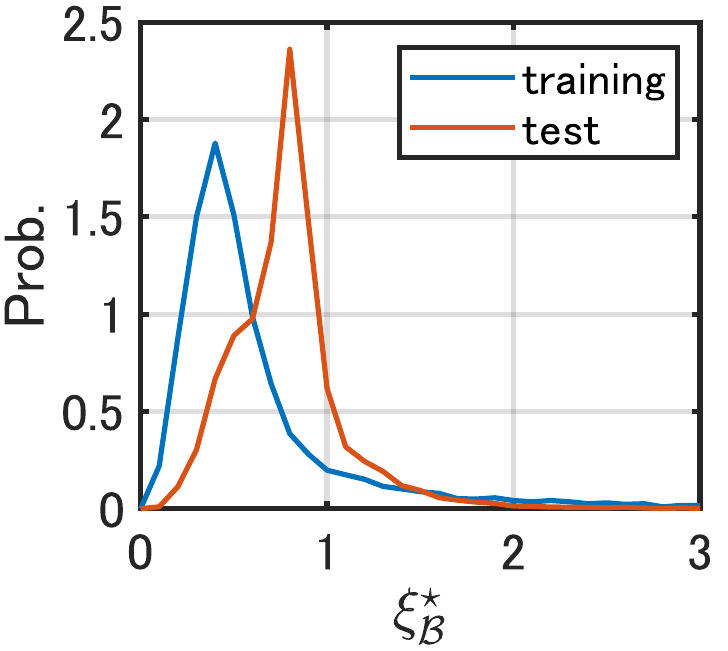}\\
    (a)  & (b)
    \end{tabular}
    }
    \end{center}
}
    \vspace{-2mm}
    \caption{Mini-batch statistics. (a) The number of mini-batches is counted up in the $i$-th box given by $\mathrm{arg\:max}_i | v_i^\top \partial \mathcal{L}_\mathcal{B} / \partial \theta |_{\theta=\theta_0} |$, where $v_i$ represents the Hessian eigenvector of the $i$-th eigenvalue ($i=1$ being the largest). The mini-batch gradients most frequently show highest correlations to the principal eigenvector. (b) Probability distributions of $\xi_\mathcal{B}^\star$ (see Eq.~(\ref{eq_xistar}) for definition) are evaluated. Positions of the mini-batch loss minima for the training set tend to appear smaller than those for the test set. These experiments use a machine-generated toy dataset with an MLP with 5 weight layers.}
    \label{fig_eigenmode}
\end{figure}

PoF determines the direction of perturbation in a data-driven manner.
Perturbing in a spherically uniform way would be very inefficient, because Hessian spectra are in most cases dominated by a very small number of eigenstates compared to the parameter dimension, and the rest of the Hessian components are negligible.
Thus, perturbing along the direction of the eigenvector corresponding to the maximum eigenvalue of the Hessian would be much more efficient.
However, it is practically infeasible to compute a Hessian and its eigenvectors even at some interval of iterations due to high computational cost.
PoF avoids to directly compute Hessians and adopts a much more computationally efficient approach.
Figure~\ref{fig_eigenmode}~(a) shows counts of randomly sampled mini-batches whose gradients maximally correlate with $i$-th eigenvector of the Hessian.
This toy experiment shows that mini-batch gradients highly likely correlate the principal eigenvector.
Though we are unsure to what extent this tendency holds, we simply assume this tendency generally holds.
Based on the observation, we adopt an approach where a perturbation is taken along a mini-batch gradient direction in the classifier parameter space.

PoF determines the range of perturbation in a data-driven manner as well.
It adopts a linear search method along the direction of (negative) mini-batch gradients evaluated at $\theta_0$ to find the nearest minimum in the 1D subspace, \ie,
\begin{equation}
    \xi_\mathcal{B}^\star = \argmin_{\xi\ge 0} \mathcal{L}_\mathcal{B} \left(\phi, \theta_0 - \xi \hat{\mathcal{L}_\mathcal{B}'} \right),
    \label{eq_xistar}
\end{equation}
where $\hat{\mathcal{L}_\mathcal{B}'}$ is a special short-hand notation of
\begin{equation}
    \hat{\mathcal{L}_\mathcal{B}'} \equiv \frac{\partial\mathcal{L}_\mathcal{B} / \partial\theta|_{\theta = \theta_0}}{|| \partial\mathcal{L}_\mathcal{B} / \partial\theta|_{\theta = \theta_0} ||_2}.
\end{equation}
Here, $\xi_\mathcal{B}^\star$ is the Euclidean distance to the minimum of the mini-batch loss in the 1D subspace.
The classifier with the perturbed parameters $\theta_0 - \xi \hat{\mathcal{L}_\mathcal{B}'}$ behaves as a somewhat weak classifier for the entire training dataset when $\xi \simeq \xi_\mathcal{B}^\star$.
The idea here is to make this kind of a somewhat weak classifier stronger for an arbitrary mini-batch $\mathcal{B}$ by optimizing the feature extractor.
In this way, the perturbed region is expected to become a flat basin.
Next, we investigate an appropriate perturbation range of $\xi$.
Figure~\ref{fig_eigenmode}~(b) shows probability distributions of $\xi_\mathcal{B}^\star$ for different mini-batches after an orthodox SGD training using a toy dataset.
It indicates that the peak point of the test distribution is roughly twice as that of the training distribution.
Then, a naive strategy would be to enlarge the perturbation range by setting $\xi = \gamma \xi_\mathcal{B}^\star$ with $\gamma \simeq 2$ to compensate the training-test distribution gap.

One could use an iterative gradient descent method instead of linear search to find a nearest minimum of a mini-batch loss.
But, the computational cost would become much higher in such a case.
Given that the classifier parameters $\theta_0$ already reached a local minimum, it would be a decent assumption that a mini-batch loss landscape is locally convex.
Computationally efficient linear search usually suffices in practice.

\begin{equation}
    \phi^\star = 
    \argmin_\phi
    \frac{1}{m}
    \sum_{\mathcal{B} \in \{ \mathcal{B}_1, \mathcal{B}_2, \cdots, \mathcal{B}_m \}}
    \mathcal{L}_\mathcal{D} 
    (\phi, \theta_0 - \gamma \xi_\mathcal{B}^\star \hat{\mathcal{L}_\mathcal{B}'})
    \label{eq_grandloss}
\end{equation}
for some $m \gg 1$.
An update of feature-extractor parameters, $\Delta\phi$, that approximately minimizes Eq.~(\ref{eq_grandloss}), is given by
\begin{equation}
    \Delta \phi = - \eta \frac{\partial \mathcal{L}_{\tilde{\mathcal{B}}} (\phi, \theta_0 - \gamma\xi_\mathcal{B}^\star \hat{\mathcal{L}'_\mathcal{B}})}{\partial \phi}, ~~
    \eta > 0
    \label{eq_DeltaPhi}
\end{equation}
with arbitrarily chosen mini-batches $\mathcal{B}$ and $\tilde{\mathcal{B}}$.
Here, we artificially set $\partial \hat{\mathcal{L}'_\mathcal{B}} / \partial \phi \triangleq 0$ so that the update reduces the empirical loss with respect to the perturbed classifier $\theta_0 - \gamma\xi_\mathcal{B}^\star \hat{\mathcal{L}'_\mathcal{B}}$.
Modification of Eq.~(\ref{eq_DeltaPhi}) by adding a momentum term, \etc, is possible.
This update makes the perturbed classifier stronger, contributing to make the local loss landscape flatter, as illustrated in Fig.~\ref{fig_algorithm}.
When $\gamma=0$, the update rule is equivalent to SGD.
As an extension of the method, one could randomize $\gamma$ in a predefined range such as $\gamma\in [0, 2]$, or employ some scheduling function to $\gamma$ such as linear growth from 0 to 2.
A pseudocode of our algorithm is provided in Algorithm~\ref{alg1}.
One trick that we employ here is that a mini-batch is re-sampled right before computing the feature-extractor update in Eq.~(\ref{eq_DeltaPhi}).
This avoids too much over-fitting to a particular mini-batch at each iteration.

PoF updates only feature extractors, and the classifier parameters $\theta_0$ are kept unchanged.
From our experience, this strategy works well on real datasets.
But, it could be possible that the position of the minimum drifts from $\theta_0$ during PoF.
To avoid this type of drifting, one may add an SGD update step for the classifier parameters $\theta_0$ after the feature-extractor parameter update.
From our experience, this strategy does not affect the final performance much on real datasets, so we did not apply this strategy for the experiments reported in the experimental section (Sec.~\ref{sec_exp}).

A similar concept to PoF was proposed by Sato~\etal\hspace{0.2mm}, where they proposed a supervised representation learning method, FOCA, for optimizing a feature extractor with respect to weak-classifier ensemble~\cite{paper:foca, paper:foca2}.
To obtain a weak classifier, FOCA applies gradient descent iterations using a given mini-batch from random initialization, while PoF effectively finds a nearest local minimum of the mini-batch loss starting from the training-loss minimum.
Not only does PoF enable post-training, but PoF can effectively find a flatter minimum, as is explained in the next subsection.

As for word choice, we adopt ``post-training" rather than ``fine-tuning" throughout this paper.
The latter word is commonly used in transfer learning settings, whereas PoF intends to improve in-distribution performance.

\subsection{Mathematical Analysis}

Next, we present a mathematical analysis about the relationship of the proposed algorithm and the loss landscape.
Let us approximate the mini-batch loss landscape along a direction of mini-batch loss gradients $\hat{\mathcal{L}_\mathcal{B}'}$ as
\begin{equation}
    \mathcal{L}_\mathcal{B} (\phi, \theta_0 - \xi \hat{\mathcal{L}_\mathcal{B}'}) \simeq \frac{1}{2} (\xi - \xi_\mathcal{B}^\star)^2 \hat{\mathcal{L}_\mathcal{B}'}^\top \mathcal{H}_\mathcal{B}(\phi, \theta_0)  \hat{\mathcal{L}_\mathcal{B}'},
    \label{eq_batchloss}
\end{equation}
where $\mathcal{H}_\mathcal{B}(\phi, \theta_0)$ is the second-order derivatives of the mini-batch loss defined in the classifier parameter space,
\begin{equation}
    (\mathcal{H}_\mathcal{B}(\phi, \theta_0))_{ij} = \left. \frac{\partial^2 \mathcal{L}_\mathcal{B}(\phi, \theta)}{\partial \theta_i \partial \theta_j} \right|_{\theta = \theta_0}.
\end{equation}
Here, we model the landscape as a quadratic function of $\xi\in\mathbb{R}$.
As is obvious from Eq.~(\ref{eq_batchloss}), $\xi = \xi_\mathcal{B}^\star$ minimizes the mini-batch loss in the linear subspace whose basis is given by $\hat{\mathcal{L}_\mathcal{B}'}$. 
It is also assumed that the minimum is zero, which is not a bad approximation, given that $\theta_0$ already minimizes the entire training loss and the mini-batch loss is further optimized by linear search.
Next, we expand the loss function of the entire training set $\mathcal{D}$ as
\begin{equation}
    \mathcal{L}_\mathcal{D} (\phi, \theta_0 - \xi \hat{\mathcal{L}_\mathcal{B}'}) \simeq \mathcal{L}_\mathcal{D} (\phi, \theta_0) + \frac{\xi^2}{2} \hat{\mathcal{L}_\mathcal{B}'}^\top \mathcal{H}_\mathcal{D}(\phi, \theta_0) \hat{\mathcal{L}_\mathcal{B}'},
    \label{eq_trainingloss}
\end{equation}
where $\mathcal{H}_\mathcal{D}(\phi, \theta_0)$ is the Hessian computed from the entire training set in the classifier parameter space.
Since the training loss is assumed to be locally minimized at $\theta = \theta_0$, there is no first-order term of $\xi$ in Eq.~(\ref{eq_trainingloss}).
Setting $\xi = \gamma \xi_\mathcal{B}^\star$, Eq.~(\ref{eq_trainingloss}) becomes
\begin{equation}
    \mathcal{L}_\mathcal{D} (\phi, \theta_0 - \gamma \xi_\mathcal{B}^\star \hat{\mathcal{L}_\mathcal{B}'}) \simeq \mathcal{L}_\mathcal{D} (\phi, \theta_0) + \frac{\gamma^2 \xi_\mathcal{B}^{\star 2}}{2} \hat{\mathcal{L}_\mathcal{B}'}^\top \mathcal{H}_\mathcal{D}(\phi, \theta_0)  \hat{\mathcal{L}_\mathcal{B}'}.
    \label{eq_effective1}
\end{equation}
Let us call this approximated loss as the effective loss.
The algorithm effectively reduces these quantities for arbitrary choices of $\mathcal{B}$ by searching appropriate feature-extractor parameters $\phi$, on which $\mathcal{L}_\mathcal{D}$ and $\mathcal{H}_\mathcal{D}$ depend.
The effective loss consists of the zero-th and the second order derivatives of the training loss.
If $\xi_\mathcal{B}^\star$ goes small, the zero-th order term becomes a dominant term; oppositely, if $\xi_\mathcal{B}^\star$ goes large, the second order term becomes a dominant term.
The quantity
$\frac{\gamma^2 \xi_\mathcal{B}^{\star 2}}{2} \hat{\mathcal{L}_\mathcal{B}'}^\top \mathcal{H}_\mathcal{D} \hat{\mathcal{L}_\mathcal{B}'}$
is a rough estimate of $\Delta\mathcal{L}$ as depicted in Fig.~\ref{fig_explanation} for a subset of the test set.
Reduction of this curvature-based loss term would decrease the training-loss curvature as well as the test-loss curvature, under the assumption that two curvatures are interlocked. 
By combining Eq.~(\ref{eq_batchloss}) and Eq.~(\ref{eq_trainingloss}), Eq.~(\ref{eq_effective1}) can be equivalently expressed as
\begin{equation}
        \mathcal{L}_\mathcal{D} (\phi, \theta_0 - \gamma \xi_\mathcal{B}^\star \hat{\mathcal{L}_\mathcal{B}'}) \simeq \mathcal{L}_\mathcal{D} (\phi, \theta_0) + \gamma^2 \mathcal{L}_\mathcal{B} \frac{\hat{\mathcal{L}_\mathcal{B}'}^\top \mathcal{H}_\mathcal{D} (\phi, \theta_0) \hat{\mathcal{L}_\mathcal{B}'}}{\hat{\mathcal{L}_\mathcal{B}'}^\top \mathcal{H}_\mathcal{B}  \hat{\mathcal{L}_\mathcal{B}'}}.
    \label{eq_effective2}
\end{equation}
The second order term has the batch-gradient projected Hessian component with a coefficient given by batch statistics $\mathcal{L}_\mathcal{B}$, $\hat{\mathcal{L}_\mathcal{B}'}$, and $\mathcal{H}_\mathcal{B}$ evaluated at $(\phi, \theta_0)$ besides the overall scalar $\gamma$.
In this way, the range of perturbation is determined in a data-driven manner.
As described earlier, $\gamma$ has an intuitive geometrical meaning, such that $\gamma=1$ corresponds to the minimum of the mini-batch loss and $\gamma=2$ corresponds to the opposite side of the quadratic mini-batch loss landscape.

It is worth mentioning that the proposed method often gives some order-of-magnitude large perturbations in the classifier parameter space compared to a typical SGD step.
Nevertheless, the algorithm is surprisingly stable.
Suppose now that $\gamma = 2$.
If the mini-batch loss is quadratic as in Eq.~(\ref{eq_batchloss}), a parameter perturbation with $\gamma=2$ makes the mini-batch loss invariant.
It means this mini-batch experiences no harm by this (possibly very large) perturbation.
This prevents the classifier from becoming too adversarial to the rest of the samples.

In reality, some mini-batches show asymmetric loss landscapes, such that one side for $\xi\in (0, \xi_\mathcal{B}^\star)$ is close to quadratic but the other side is almost constant.
This type of loss landscapes can be detected in the linear search.
When such a landscape is found during the training, one may discard the mini-batch and repeat the mini-batch sampling step followed by the linear search algorithm.

\section{Experiments}
\label{sec_exp}

The aim of this section is to show quantitative results about generalization performance gain, the change in target Hessian components, training time, scale of perturbation ranges by PoF, and further classifier post-training.
We conducted various image classification experiments on CIFAR-10, CIFAR-100~\cite{Krizhevsky09learningmultiple}, SVHN~\cite{SVHN}, and Fashion-MNIST \cite{paper:fmnist}. 
WideResNet-28-10 \cite{paper:wideresnet} was used as the classification network in all experiments. 

\subsection{Settings}

\noindent {\bf Baseline methods.} 
We compared the performance of PoF to SGD and SAM as baselines.
The network was trained for 250 epochs with batch size of 256. 
The learning rate was initialized to 0.1 (0.01 for SVHN) and was multiplied by a factor of 0.2 at 60-th, 120-th, 160-th, and 200-th epochs.\footnote{We also tried different learning rate, namely 3e-5, after 200-th epoch as adopted for PoF.
The resulting test error rates of SAM are similar to the corresponding values in Table~\ref{tab:comparison}.}
We used the Nesterov Accelerated Gradient~\cite{nesterov1998introductory} with momentum rate of 0.9 and weight decay rate of 5e-4. 
With SAM, $\rho$, the range of the perturbation, was set to 0.05 (0.01 for SVHN) as in the original paper.\footnote{For the record, we also tested SAM with two order-of-magnitude larger value of $\rho$ on CIFAR-10, and found that the test accuracy is similar to that with the default $\rho (=0.05)$.} 
Weights in the feature extractors use He-initialization, and those in classifiers were initialized with a normal distribution $\mathcal{N}(0,0.1^2)$.

\noindent {\bf Training details of PoF.} The network was trained with SAM ($\rho=0.05$) for the first 200 epochs. Then, the feature extractor was post-trained with PoF for additional 50 epochs with batch size of 256 and learning rate of 3e-5, with the Nestrov Accelerated Gradient having the same parameters with those in SGD. 
The batch size for generating weak classifiers was 32. 
From our experience, this batch size was sufficient for PoF to work well. The expansion factor $\gamma$ in Eq.~(\ref{eq_DeltaPhi}) was randomly sampled at each iteration from a predefined range, \ie, $\gamma\in [0, 2]$ in all experiments.

\noindent{\bf Additional details.} 
All results used basic data augmentations (horizontal flip, padding by four pixels, and random crop), and cutout with $16\times 16$ pixels was additionally used for the results of CIFAR-\{10, 100\}. 
We used standard training/validation/testing split for all datasets, but the 530K extra images were used in addition to the standard training data of SVHN.
The computing environment used in all experiments is 4 compute nodes, each equipped with 4 NVIDIA A100 GPUs, \ie, totally 16 GPUs were used in parallel.

\subsection{Results}

\begin{figure}[t]
{\normalsize
    \begin{center}
    {\tabcolsep=1mm
    \begin{tabular}{cc}
    \includegraphics[height=39mm]{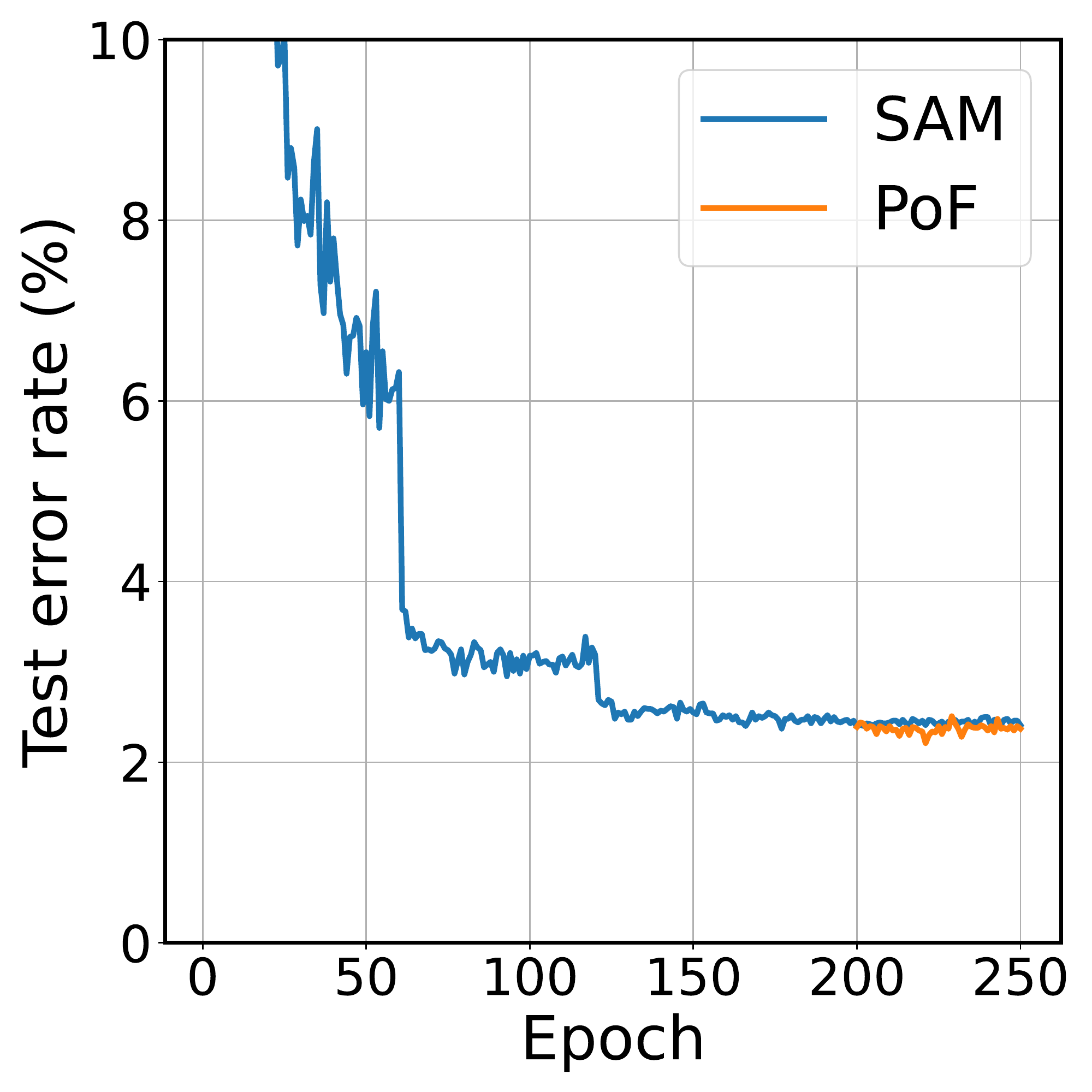}& 
    \hspace{0mm}
    \includegraphics[height=39mm]{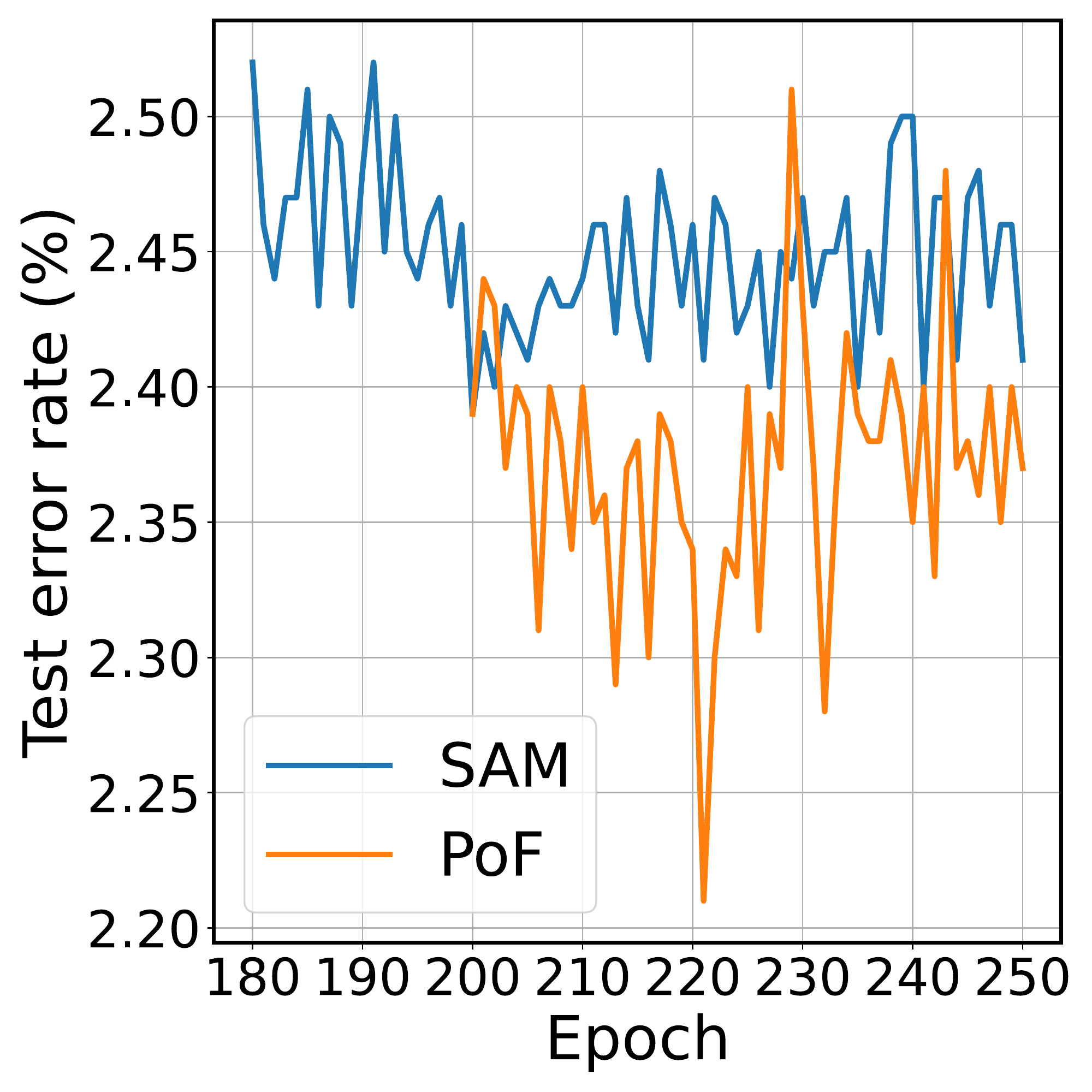}\\
    \small (a) Entire test error rate curves  & \small (b) Zoomed view
    \end{tabular}
    }
    \end{center}
    \vspace{-2mm}
}
    \caption{The test error rate curves of SAM and PoF. (a) shows the behavior of overall training duration. (b) provides an enlarged view between 180-250 epochs. PoF post-trains the network that is trained by SAM for 200 epochs. For a comparison, we also shows the plot of SAM that is continued after 200 epochs. PoF enjoys further error rate drop right after it gets activated.}
    \label{fig_test_error_epoch}
\end{figure}

\noindent{\bf Generalization.} The test error rates with SGD, SAM, and PoF with different training epochs are summarized in Table~\ref{tab:comparison}. 
Two checkpoints were used for each method; namely, SGD at 200/250 epochs, SAM at 200/250 epochs, PoF at 210 epochs (10-epoch post-trained), PoF at 250 epochs (50-epoch post-trained).
As the table shows, on CIFAR-\{10, 100\} and SVHN, PoF can improve the classification accuracy on average.
In each case, the performance gain of averaged accuracy from the second-to-the-best result clearly exceeds by one standard deviation.
The timing of PoF's peak performance somewhat varies depending on datasets.
On CIFAR-\{10, 100\}, the peak performance comes relatively early, say about 10-20 epochs after PoF activated, as is shown in Fig.~\ref{fig_test_error_epoch}.
On SVHN it comes relatively later, say by 40 epochs.
But in either case, training epochs can be much fewer, compared to training from random initialization.

The result shows that PoF does not improve generalization for Fashion-MNIST.
This indicates that flattening the loss landscape in the classifier parameter space could not further improve the performance, probably because lower layers suffers severer loss deterioration $\Delta\mathcal{L}$. 
PoF clearly has a limitation in such a case.

\begin{table}[t]
\caption{Test error rates (\%) of classification on CIFAR-\{10, 100\}, SVHN, and Fashion-MNIST. Two check points are evaluated for each method. For PoF, networks that are trained by SAM for the first 200 epochs are post-trained for additional 10 epochs (totally 210 epochs as depicted) and for additional 40 epochs (totally 250 epochs as depicted). 
PoF outperforms three out of four datasets.}
\label{tab:comparison}
\vspace{2mm}
\scalebox{0.8}[0.8]{
\begin{tabular}{c|cccc|}
\cline{2-5}
\multicolumn{1}{l|}{}                                                            & \multicolumn{4}{c|}{Dataset}                                                                         \\ \hline
\multicolumn{1}{|c|}{Method}                                                     & \multicolumn{1}{c|}{CIFAR-10} & \multicolumn{1}{c|}{CIFAR-100} & \multicolumn{1}{c|}{SVHN} & Fashion \\ \hline
\multicolumn{1}{|c|}{\begin{tabular}[c]{@{}c@{}}SGD\\ (200 epochs)\end{tabular}} & \multicolumn{1}{c|}{3.22$\pm$0.14} & \multicolumn{1}{c|}{18.23$\pm$0.35} & \multicolumn{1}{c|}{1.67$\pm$0.03} & 4.60$\pm$0.11    \\ \hline
\multicolumn{1}{|c|}{\begin{tabular}[c]{@{}c@{}}SGD\\ (250 epochs)\end{tabular}} & \multicolumn{1}{c|}{3.14$\pm$0.13} & \multicolumn{1}{c|}{18.40$\pm$0.35} & \multicolumn{1}{c|}{1.67$\pm$0.03} & 4.63$\pm$0.14    \\ \hline
\multicolumn{1}{|c|}{\begin{tabular}[c]{@{}c@{}}SAM\\ (200 epochs)\end{tabular}} & \multicolumn{1}{c|}{2.50$\pm$0.07} & \multicolumn{1}{c|}{16.27$\pm$0.09} & \multicolumn{1}{c|}{1.64$\pm$0.04} &    4.14$\pm$0.09     \\ \hline
\multicolumn{1}{|c|}{\begin{tabular}[c]{@{}c@{}}SAM\\ (250 epochs)\end{tabular}} & \multicolumn{1}{c|}{2.53$\pm$0.08} & \multicolumn{1}{c|}{16.32$\pm$0.20} & \multicolumn{1}{c|}{1.63$\pm$0.03} & \textbf{4.12}$\pm$0.05     \\ \hline \hline
\multicolumn{1}{|c|}{\begin{tabular}[c]{@{}c@{}}SAM$\shortrightarrow$PoF \\ (210 epochs)\end{tabular}} & \multicolumn{1}{c|}{\textbf{2.41}$\pm$0.02} & \multicolumn{1}{c|}{\textbf{16.07}$\pm$0.15} & \multicolumn{1}{c|}{1.60$\pm$0.04} & 4.25$\pm$0.05      \\
\multicolumn{1}{|c|}{\begin{tabular}[c]{@{}c@{}}SAM$\shortrightarrow$PoF \\ (250 epochs)\end{tabular}} &
\multicolumn{1}{c|}{\textbf{2.41}$\pm$0.06} & \multicolumn{1}{c|}{16.60$\pm$0.05} & \multicolumn{1}{c|}{\textbf{1.55}$\pm$0.02} &   4.35$\pm$0.07      \\ \hline
\end{tabular}
}
\vspace{4mm}
\end{table}

\begin{figure}[t]
{\normalsize
    \begin{center}
    {\tabcolsep=1mm
    \begin{tabular}{cc}
    \includegraphics[height=39mm]{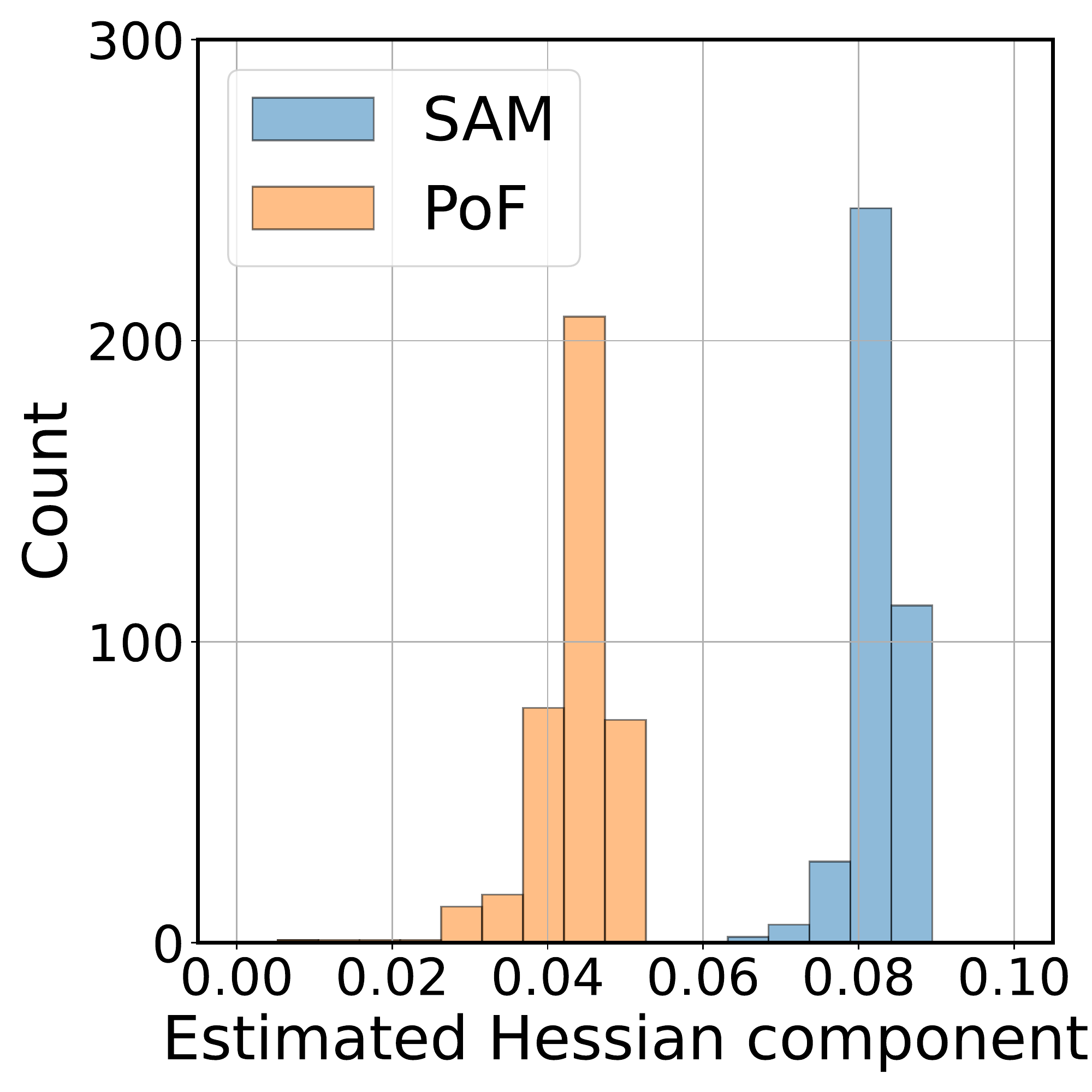} 
     & 
    \hspace{0mm}
    \includegraphics[height=39mm]{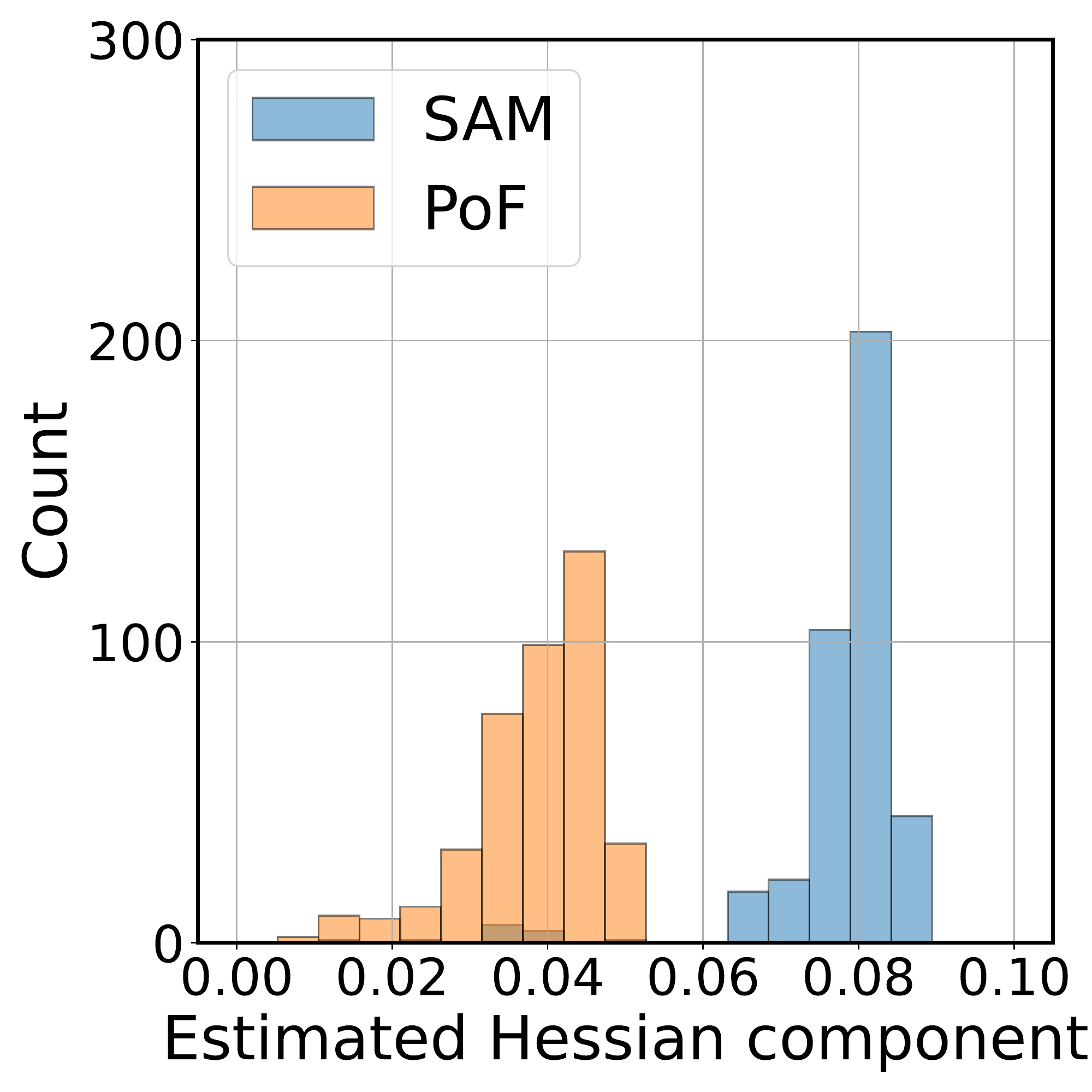}\\
    \small (a) Training dataset & \small (b) Test dataset
    \end{tabular}
    }
    \vspace{-3mm}
    \end{center}
}
    \caption{Histograms of estimated Hessian components on (a) training dataset and (b) test dataset. Blue bars show results of SAM at 200 epochs and orange bars show results of PoF at 210 epochs (10-epoch post-trained). PoF successfully reduces Hessian components from SAM.  See text for the way of estimation.}
    \label{fig_hessian_freq}
\end{figure}

\noindent{\bf Hessian evaluation.}
We evaluated how particular Hessian components at the classifier parameter space change by PoF.
Since direct computations of Hessian matrices are computationally demanding, we adopt a computationally efficient way of estimating the largest eigenvalue of the Hessian block.
As discussed in Sec.~\ref{sec_intro}, a mini-batch gradient shows high correlation to the principal eigenvector of Hessian matrix.
We gathered 400 such estimations with different mini-batches on CIFAR-10, and made histograms as shown in Fig.~\ref{fig_hessian_freq}.\footnote{For the test distribution, mini-batch gradients are computed using test samples.}

As is evident from Fig.~\ref{fig_hessian_freq}, PoF clearly reduces the Hessian components.
The peak values are roughly reduced by a factor of two for both training and test sets.
As SAM is a strong baseline having a flatness-enhancing functionality, it is surprising that there is still a room for the network to improve flatness along certain directions as well as to improve generalization just by additional 10 epoch post-training.

\begin{table}[t]
    \begin{center}
     \caption{Training time per epoch. WideResNet-28-10 was used on the CIFAR-10 dataset. Data-parallelism is utilized with 16 NVIDIA A100 GPUs.}
     \label{tab:time}
     \vspace{2mm}
    \label{fig_training_time}
    \begin{tabular}{|c|c|}
    \hline
    Method & Time per epoch \\
    \hline
    SGD & 21.8 s\\ \hline
    SAM & 32.9 s \\ \hline
    PoF & 25.6 s \\ \hline
    \end{tabular}
    \end{center}
\end{table}


\begin{figure}[t]
{\normalsize
    \begin{center}
    \includegraphics[height=50mm]{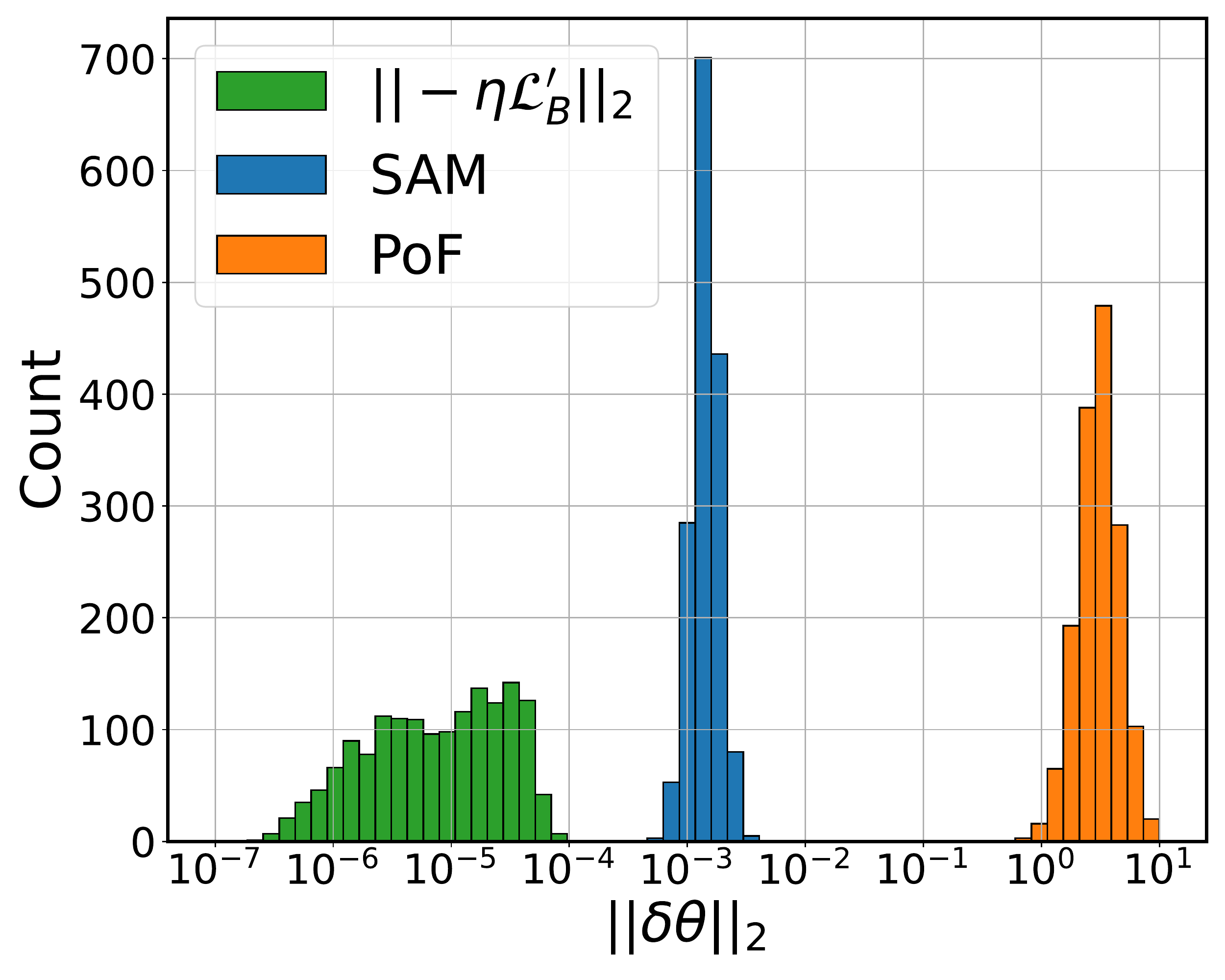}
    \end{center}
}
    \caption{Histograms of scales of perturbation range. The $x$-axis shows sizes of classifier-parameter perturbations denoted by $||\delta \theta||_2$ in a logarithmic scale. PoF exhibits a few order-of-magnitude larger perturbations than SAM, while SAM does have large perturbations compared to typical size of SGD updates.}
    \label{fig_perturb_range}
\end{figure}

\noindent{\bf Computational time.}
In Table \ref{tab:time}, we show the comparison of training time per epoch for SGD, SAM and PoF. 
SAM requires $1.51\times$ more time than SGD, and PoF requires $1.17\times$ more time than SGD per epoch. 
SAM requires computation of gradients multiple times to generate a parameter perturbation.
This additional process slows the training speed.
In contrast, PoF adopts a simple linear search algorithm; thus, only a single gradient computation is required to generate parameter perturbation.

\noindent{\bf Perturbation range.} 
We compared the scale of the parameter perturbations in the classifier parameter space $\theta$ to those of SGD and SAM. 
\underline{\textit{PoF setting}}:
The perturbation sizes are given by $||-\gamma \xi_\mathcal{B}^\star \hat{\mathcal{L}_\mathcal{B}'}||_2$ (see Eq.~(\ref{eq_grandloss})) for different mini-batches $\mathcal{B}$, and we set $\gamma=2.0$.
One of the CIFAR-10 models post-trained by PoF for 10 epochs (after pre-training by SAM for 200 epochs) was used.
\underline{\textit{SGD setting}}:
As SGD is not a perturbation-based method, we simply measured sizes of regular updates for reference given as $||-\eta \nabla \mathcal{L}_\mathcal{B}||_2$ for different $\mathcal{B}$.
The learning rate $\eta$ was set to 1.6e-4 to measure typical update sizes. 
One of the CIFAR-10 models trained by SGD for 200 epoch was used.
\underline{\textit{SAM setting}}:
Sizes of a classifier-parameter perturbations, which are upper bounded by the fixed radius $\rho$, were measured for different $\mathcal{B}$. 
One of the CIFAR-10 models trained by SAM for 200 epoch was used.
\underline{\textit{Results}}:
The histograms of those scalar values of each method are shown in Fig.~\ref{fig_perturb_range}. 
The horizontal axis is shown in the logarithmic scale. A typical perturbation range involved in PoF is a few order-of-magnitude larger than that in SAM, while SAM has much larger scale than SGD updates.
PoF can have a very large perturbation range, which effectively works as expanding the flat region.
In spite of such large displacements, learning is quite stable thanks to the fact that the perturbed classifier does work well on a certain training mini-batch.

\noindent{\bf Further classifier fine-tuning.}
In all experiments shown in Table~\ref{tab:comparison}, parameter set $(\phi^\star, \theta_0)$ were used for evaluation, where $\phi^\star$ is given by PoF and $\theta_0$ is given by the pre-training method, \ie, SAM.
It means that PoF did not change the classifier parameters $\theta_0$ after all.
As discussed in the previous section, it might be possible that the position of the minimum drifts away from $\theta_0$ during PoF.
We examined this possibility by fine-tuning only $\theta_0$ with respect to fixed feature extractor $\phi^\star$ \textit{after} PoF.
We took a particular training instance from CIFAR-10 experiments.
Its test error rate after PoF at 210 epochs (10-epoch post-trained) is marked 2.40\%.
Then, starting from this model, we fine-tuned the classifier for additional 10 epochs.
Final test error rate became $2.39\pm0.03\%$.
This experiment indicates that further classifier fine-tuning does not improve performance.


\section{Conclusion}

This paper introduced PoF: Post-Training of Feature Extractor.
PoF is an in-domain post-training method that updates a feature-extractor part of a deep network that has already optimized by some method.
Motivated by a toy-data observation, we made an assumption that flattening loss landscape in the higher layer parameter space likely improves generalization, analogous to classical maximum margin methods.
Aiming to reduce large eigenvalues of Hessian defined in the higher-layer classifier parameter space,
PoF applies parameter perturbations to the classifier parameters in a particular way that reduces a curvature-aware effective loss, 
and updates the feature-extractor parameters.
It is demonstrated that PoF further improved test performance of networks that are already trained by SAM on three out of four datasets.
Notably, on certain datasets, performance improvements with clear margins were obtained by only additional 10-epoch post-training.

\section*{Acknowledgements}

This work is an outcome of a research project, Development of Quality Foundation for Machine-Learning Applications, supported by DENSO IT LAB Recognition and Learning Algorithm Collaborative Research Chair (Tokyo Tech.). Guoqing Liu provided us implementation support. Kohta Ishikawa and Teppei Suzuki gave us insightful comments.

\bibliography{PoF_ICML2022}
\bibliographystyle{icml2022}

\end{document}